\documentclass[a4paper,conference]{IEEEtran}

\usepackage{tabularx}
    \newcolumntype{L}{>{\raggedright\arraybackslash}X}
\usepackage{multirow}

\usepackage[noend]{algpseudocode}
\usepackage{algorithm}

\usepackage{arabtex}
\usepackage{utf8}

\usepackage{cite}

\usepackage{float}

\usepackage[pdftex]{graphicx}

\usepackage[symbol]{footmisc}

\usepackage[flushleft]{threeparttable}

\begin{document}
\setcode{utf8}

\title{An Efficient Language-Independent Multi-Font OCR for Arabic Script}

\author{

\IEEEauthorblockN
{Hussein Osman\IEEEauthorrefmark{1},
Karim Zaghw\IEEEauthorrefmark{1},
Mostafa Hazem\IEEEauthorrefmark{1}, and Seifeldin Elsehely\IEEEauthorrefmark{1}}

\IEEEauthorblockA{\IEEEauthorrefmark{1} Computer Engineering Department, Faculty of Engineering, Cairo University. \\ hussein.fadl@eng.cu.edu.eg, karim@zaghw.com, mostafahazem98@hotmail.com, seif.elsehely@gmail.com}
}

\maketitle
\begin{abstract}
Optical Character Recognition (OCR) is the process of extracting digitized text from images of scanned documents. While OCR systems have already matured in many languages, they still have shortcomings in cursive languages with overlapping letters such as the Arabic language. This paper proposes a complete Arabic OCR system that takes a scanned image of Arabic Naskh script as an input and generates a corresponding digital document. Our Arabic OCR system consists of the following modules: Pre-processing, Word-level Feature Extraction, Character Segmentation, Character Recognition, and Post-processing. This paper also proposes an improved font-independent character segmentation algorithm that outperforms the state-of-the-art segmentation algorithms. Lastly, the paper proposes a neural network model for the character recognition task. The system has experimented on several open Arabic corpora datasets with an average character segmentation accuracy 98.06\%, character recognition accuracy 99.89\%, and overall system accuracy 97.94\% achieving outstanding results compared to the state-of-the-art Arabic OCR systems. 

\end{abstract}
\IEEEpeerreviewmaketitle
\section{Introduction}
The problem of Optical Character Recognition (OCR) has been the scope of research for many years \cite{namysl2019efficient, dixit2018optical, goodfellow2013multi} due to the need for an efficient method to digitize printed documents, prevent their loss and gradual unavoidable wear, as well as increase their accessibility and portability. The challenges that face Arabic OCR systems stem from the cursive and continuous nature of Arabic scripts. The presence of a semi-continuous baseline in Arabic text prevents the use of segmentation techniques proposed for other OCR systems. Moreover, the vertical overlapping of characters caused by ligatures means that segmenting a word along a single horizontal line will not achieve perfect segmentation. Furthermore, the challenges introduced by the nature of the Arabic script do not only affect character segmentation. The recognition of Arabic characters requires a huge training set since each character can have a different shape depending on its position in the word. Also, cases of constant miss-classification of a character as another one are frequent due to the presence of characters that are only told apart through the number of dots.

\begin{figure*}
\centering
\includegraphics[width=\textwidth]{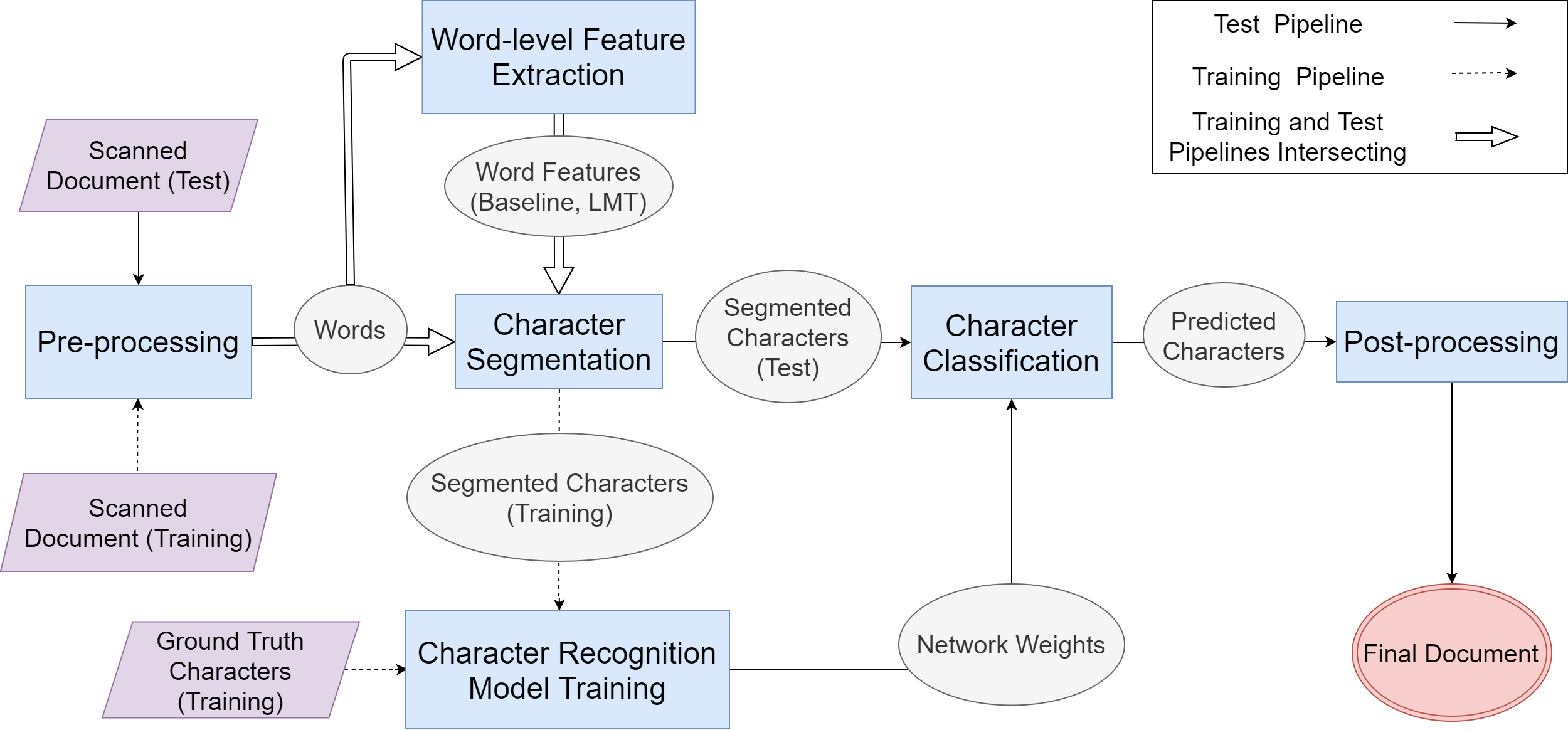}
\caption{System Architecture of our proposed workflow for optical character recognition. }
\label{SystemBlockDiagram}
\end{figure*}

In this paper, we propose a complete OCR pipeline for Arabic text that is language-independent and supports multiple fonts. Our system takes a scanned image of an Arabic document as an input and outputs a digitized text document containing the predicted text. The input image is first preprocessed where binarization, denoising, and deskewing are carried out, followed by line and word segmentation. Character segmentation is then performed based on the extracted word-level features, followed by cut filtration based on the wide rules-set we have defined. The segmented characters are then fed into our character recognition neural network model which classifies the given character into one of the 29 Arabic characters. Furthermore, post-processing is applied to the predicted characters and conveniently concatenates them into a comprehensible text document.

Our main contribution in this paper lies in the development of an efficient light-weight system that outperforms current state-of-the-art accuracy in the field of Arabic OCR. We achieve outstanding runtime results without trading off our system accuracy through efficient denoising of documents, vectorized implementation of all segmentation stages, and finely tuning our recognition model’s complexity. In addition, we maintain our overall system accuracy by improving the character segmentation using the proposed improved cut filtration algorithm. The algorithm is robust against structural, morphological, and topological similarities between letters. We also propose a neural network model to learn the underlying features of input characters and build a character classification model. We finally eliminate the use of any lexical analysis to maintain the language independence of the system.

The rest of the paper is organized as follows: Section II describes the state-of-the-art related work. Section III discusses in detail the proposed OCR system. Section IV describes the datasets used for training the neural network and for evaluating the overall system performance, then discusses the results with comprehensive comparisons with other related algorithms and methods. Section V discusses limitations in our proposed system and proposals on future work in this area.  Finally, Section VI presents the paper’s conclusion.
\section{Related Work}
There has been a variety of techniques proposed in the area of OCR for Arabic text. In this section, we will review the different approaches in the literature for character segmentation, feature extraction, and character recognition in OCR systems.

A significant number of approaches for the Arabic character segmentation task have been proposed in the literature. Mohamed \textit{et al.} \cite{mohammad2019contour} proposed the use of contour extraction to facilitate the character segmentation phase followed by cut index identification based on the contour of a word. Their method achieved significant results in character segmentation; however, its performance degraded in the case of small-sized Arabic fonts or noisy documents. The scale-space method utilized by El Makhfi \textit{et al.} \cite{el2016scale} represents another direction in Arabic character segmentation. This method works well for scanned documents containing high random noise since blobs are retrieved from characters and are then detected to recover the appropriate cut positions in the image. Although this approach has been used in several computer vision applications since its first proposal, its use in Arabic character segmentation is still widely unexplored.

Inspired by NLP applications, Alkhateeb \textit{et al.} \cite{alkhateeb2011offline} and Radwan \textit{et al.} \cite{radwan2018neural} proposed the use of a sliding window approach for segmentation. A Convolutional Neural Network (CNN) is used to determine the likelihood of a given sliding window consisting of several segments to be an actual character. Subsequently, the segments that qualify as characters are fed into their recognition model. This segmentation approach has shown very high performance on a single font but failed to maintain this high performance when tested on multiple fonts and font sizes. Lots of efforts \cite{qaroush2019efficient, zheng2004new, amara2016new} in Arabic character segmentation have been based on histogram analysis of words. 

It is important to mention that Qaroush \textit{et al.} \cite{qaroush2019efficient} also proposes a very effective word segmentation methodology that achieves state-of-the-art accuracy by implementing cut identification and filtration through gap length. Their proposed method handles multiple fonts and font sizes.

Although not being as challenging as character segmentation, feature extraction is one of the keys to boosting the accuracy of any OCR system. Rashad and Semary \cite{rashad2014isolated} adopted a simple approach that manually extracts basic features from each character such as height, width, number of black pixels, number of horizontal transitions, number of vertical transitions, and other similar features. Rashid \textit{et al.} \cite{rashid2013low} proposed a multi-dimensional Recurrent Neural Network (RNN) for their recognition system that achieved outstanding character recognition rates. However, the effects of using this complex approach on the runtime were not properly investigated. The use of deep learning approaches is highly efficient when developing Arabic OCR systems that operate on unconstrained scene text and video text, not scanned documents \cite{jain2017unconstrained}.

Dahi \textit{et al.} \cite{dahi2015primitive} adopted a similar approach by manually selecting features from a noise-free and pre-segmented character input. They added a font recognition module before the feature extraction, to include the font as a feature for the character recognition, alongside other slightly complex features such as ratios between black pixel count per region and the statistical centre of mass for each character proposed by \cite{rosenberg2012using}. The overall system of \cite{dahi2015primitive} achieved very high accuracy for Arabic character recognition.

It is worth mentioning that the OCR system of \cite{dahi2015primitive} did not include a character segmentation module as it worked only on a pre-segmented character input. Additionally, the OCR architecture of Dahi \textit{et al.} \cite{dahi2015primitive} failed to scale up and recognize Arabic characters in other fonts that were not supported by the font recognition module. A convenient middle ground between ineffective manual extraction \cite{rashad2014isolated, dahi2015primitive}, and the computationally expensive use of deep learning \cite{rashid2013low, jain2017unconstrained} is presented by the use of Principal Component Analysis (PCA) for automatic feature extraction.

As proposed by Shayegan and Aghabozorgi \cite{shayegan2014new}, PCA provides an efficient and effective solution for the problem of feature extraction in recognizing Arabic numerals. However, applying PCA becomes computationally infeasible for the huge datasets needed for training Arabic character recognizers.

As for character recognition, implementing this module without the use of machine learning has been deemed obsolete; because of the outstanding results achieved by machine learning recognition models. Hence, we will only consider the techniques for character recognition that are based on machine learning for review in the subsequent paragraphs. Shahin \cite{shahin2017printed} proposed using linear and ellipse regression to generate codes from the segmented characters. This approach of codebook generation and code matching showed average results for character recognition. Additionally, it suffered from the same problem of not being able to generalize to other Arabic fonts as \cite{dahi2015primitive}. The use of the holistic approach in \cite{nashwan2018holistic} emerged from the difficulty of the segmentation phase as we mentioned. This word-level recognition technique skips all the inaccuracy produced by segmentation errors but creates the need for post-recognition lexical analysis, thus resulting in a language-dependent system that relies on a look-up data base and semantic checks after recognition.

Often paired with the use of a sliding window for segmentation, the use of a Hidden Markov Model (HMM) for character recognition was adopted by \cite{nashwan2018holistic, rashwan2007arabic, pechwitz2003hmm, amor2006multifont}. By using a model that mimicked the architecture of an Automatic Speech Recognition system and an HMM, Rashwan \textit{et al.} \cite{rashwan2007arabic} managed to overcome the challenges presented by the presence of ligatures. Many OCR systems use other classical machine learning techniques in their character recognition module; such as random forests \cite{rashad2014isolated, dahi2015primitive}, K-Nearest Neighbor \cite{rashad2014isolated}, shallow neural networks \cite{al2006new}. The neural network model used by Al-Jarrah \textit{et al.} \cite{al2006new} yielded the best results, compared to other classical machine learning techniques \cite{rashad2014isolated, dahi2015primitive, rashwan2007arabic, amor2006multifont}, and was able to generalize over different fonts.

\section{Method Proposed}
As shown in Figure \ref{SystemBlockDiagram}, the input to the OCR system is expected to be a number of scanned images of computerized Arabic documents. In the pre-processing stage, we apply image preprocessing through the filtering, de-skewing, and de-noising of the input images, followed by line and word segmentation. In the Word-Level Feature Extraction stage, we generate statistical, structural, and topological features for every word. In the Character Segmentation stage, we apply the Excessive Cut Creation and Improved Cut Filtration algorithms to segment the word into individual characters. These segmented characters, together with the associated ground truth labels, represent the dataset that our Character Recognition Model uses for training. We train an Artificial Neural Network (ANN) to classify each segmented character into one of the 29 possible characters in the Arabic language. Finally, we aggregate the segmented characters into words and generate the output of our OCR system.
\subsection{Preprocessing}
\begin{figure}
\centering
\includegraphics[width=0.5\textwidth]{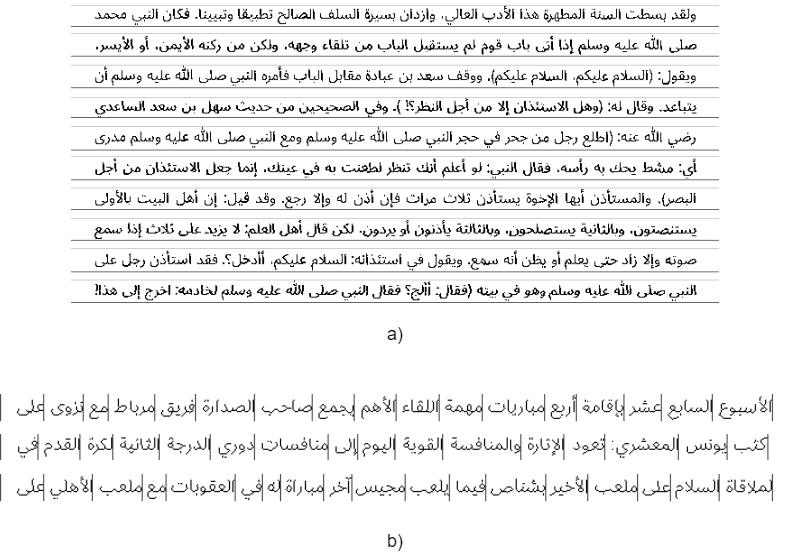}
\caption{A preprocessed text with \textbf{a)} cut indices for line segmentation \textbf{b)} cut indices for word segmentation}
\end{figure}

The preprocessing module consists of four main steps: raw image filtering, document de-skewing, line segmentation, and word segmentation. Initially, we start by converting the input image to grayscale, then binarizing it by applying Adaptive Gaussian Thresholding. The document de-skewing is carried out by rotating the document about its geometric centre with a specific angle calculated through obtaining the orientation of the text’s minimum bounding rectangle. This is followed by another round of binary thresholding to set binary values for the pixels that have been interpolated due to the previous rotation. The line segmentation step is performed by blurring the image then applying horizontal histogram projection of black pixels. Local minima of this histogram indicate positions of separations between lines. The goal of blurring is to avoid generating segmented lines containing dots only (e.g. the dots of the ‘yaa’ letter at the end of the word) or containing special Arabic diacritics only (e.g. ‘hamza’ or ‘shadda’). For word segmentation, we applied thinning to the image instead of blurring. We propose thinning as a solution to enhance the fine details within and between words. This eliminates the overlapping pixels between any two words (if one of them ends with a curved letter) and results in standardized gap lengths between words. We subsequently implemented \cite{qaroush2019efficient}’s algorithm for cut identification and filtration based on gap lengths.
\subsection{Word-level Feature Extraction}
The word-level feature extraction algorithm takes the segmented words as an input and generates for each word several geometric features. These features are essential for the character segmentation algorithm to be able to identify individual characters and segment them accordingly. We discuss the generated features for each segmented word in this subsection.

\subsubsection{Baseline} The baseline is an imaginary horizontal line that connects all of the letters in an Arabic word \cite{qaroush2019efficient}. In order to detect the baseline for every word, we search for the row of pixels with the greatest number of black pixels by applying horizontal histogram projection and finding the global maximum. 

\subsubsection{Line of Maximum Transitions (LMT)} We define a transition as a change in pixel value from 0 (black) to 1 (white) or vice versa. An important feature of the Arabic script is that a transition above the baseline is always due to a character being drawn. The Line of Maximum Transitions (LMT) is the line that cuts through the greatest number of these transitions (i.e. the row of pixels in which the number of transitions from black to white and white to black pixels is greatest)\cite{qaroush2019efficient}. For better estimation of the baseline and LMT, we propose that both features should be derived from the whole line of text rather than from each word. 

\subsubsection{Potential Cut Region (PCR)} The LMT’s key characteristic is that it passes through all potential characters in an Arabic word and is therefore essential in separating the word into its individual character components. A cut is defined as an imaginary line that separates two characters, and a Potential Cut Region is the area where a cut may exist. Since we cannot at first determine which character-intersections with the LMT belong to the same character and which are the result of a new character being written, we assume that each intersection represents a distinct character and therefore a PCR exists between any two successive intersections.

In order to determine the start and end indices of PCRs, we traverse the LMT from right to left. Each black pixel followed by a white pixel is defined as a start index of a PCR and each white pixel followed by a black pixel is defined as an end index. Furthermore, the column of pixels that is chosen as the location of a cut is called a cut index.

\begin{figure}
\centering
\includegraphics[width=2in]{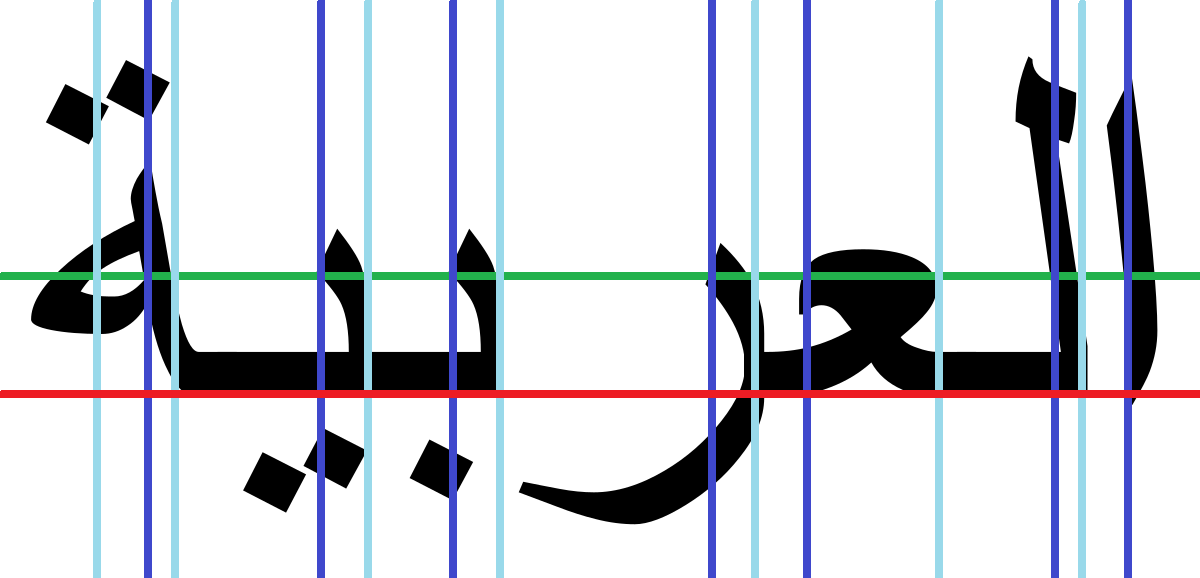}
\caption{Arabic word with the baseline highlighted in red, the LMT highlighted in green, the PCR start indices highlighted in dark blue, and the PCR end indices highlighted in light blue.}
\label{Word Features}
\end{figure}

\subsection{Character Segmentation}
To solve the problem of over-segmentation of characters, our character segmentation algorithm consists of two main steps: Excessive Cut Creation (ECC), which generates excessive potential cuts, and Improved Cut Filtration (ICF), which filters the false cuts from these potential cuts and outputs a set of valid cuts only. The Improved Cut Filtration algorithm is considered an improvement over the Cut Filtration algorithm proposed by Qaroush \textit{et al} \cite{qaroush2019efficient}.  

\subsubsection{Excessive Cut Creation Algorithm (ECC)}
In the Arabic script, characters are either connected through the baseline or separated by a single space. Therefore, there are two different methods used in the Excessive Cut Creation algorithm: Finding Baseline Cuts and Finding Separation Cuts. The former deals with separating baseline-connected characters and some space-separated characters while the latter addresses the remaining space-separated characters.

\textbf{i. Baseline Cuts (BC):} In order to identify a baseline cut, we inspect each column of pixels in a PCR, starting from the end index to the start index. We count the total number of black pixels above and below the baseline and, for every PCR, we propose the position of the cut index to be the first column where the count is zero, i.e. where the only black pixel allowed is the baseline. This approach is useful as a preliminary step for separating baseline-connected characters and also helps in separating some space-separated characters; e.g. the ‘aleph’, ‘daal’ or ‘thaal’ \<(ــا، ــد ، ــذ)> followed by another letter. 

\textbf{ii. Separation Cuts (SC):} While baseline cuts succeed in separating some space-separated characters, it will not place a cut whenever a black pixel exists below the baseline. This introduces a real challenge for letters that have curves which dip below the baseline such as ‘raa’ and ‘zeen’ \< (ـر، ـز) > since the entire PCR may contain black pixels below the baseline. To solve this problem, we place a separation cut whenever the pixels at the left and right indices of a PCR are not connected by an uninterrupted path of black pixels. This ensures that a cut will be placed wherever there are two space-separated characters even if one of them happens to dip below the baseline. We choose the separation cut index to be the middle of this PCR. 

As illustrated in Algorithm \ref{ECC_Algorithm}, we search every PCR for a baseline cut. If we find a baseline cut, then we add this cut to the set of cutIndices. If we fail to find a baseline cut, we search for a separation cut. If we find a separation cut, then we add this cut to the set of cutIndices. If we find neither a baseline cut nor a separation cut, we claim that this PCR contains a part of a character that should not be cut.

\begin{algorithm}
\begin{algorithmic}[1]

\State \textbf{Input: $PCRArray$}
\State {$CutIndices \gets \emptyset$}
\ForAll {$PCR$ in $PCR Array$}
	\State {$CutFound \gets False$}
	
	\ForAll {$PixelColumn$ in $PCR$}
		
		\If {$VProjAboveBL + VProjBelowBL = 0$}
			\State {$CutFound \gets True$}			
			\State {$CutIndices.add(PixelColumn.Index)$}
			\State \textbf{break}
		\EndIf
	\EndFor
	
	\If { \textbf{not} $CutFound$ and \textbf{not} $PCR.IsConnected$}
		\State {$CutIndices.add(PCR.MiddleIndex)$}
	\EndIf
\EndFor
\State \textbf{Output: $CutIndices$}

\end{algorithmic}
\caption{ECC Algorithm}
\label{ECC_Algorithm}
\end{algorithm}

\subsubsection{Improved Cut Filtration Algorithm (ICF)}

After generating a relatively large number of potential cut indices in the ECC stage, we begin inspecting each Potential Character (PC), where a PC is defined as any region that exists between two successive cuts. The goal of the cut filtration stage is to determine which of the cut indices are excessive false cuts. We identify the Arabic letters that usually cause false cuts in the cut filtration algorithm in the paper written by \cite{qaroush2019efficient} and other character segmentation algorithms and we propose an improved algorithm to detect all of these letters. To the best of our knowledge, there is no single algorithm that can perfectly segment all Arabic letters, and hence we introduce our solution as the new state-of-the-art. We will discuss these challenging cases, and how ICF handles each of them accordingly.

\textbf{i. ‘Seen’ and ‘Sheen’ case \<(س، ش)>}: The most notable causes of excessive false cuts are the letters ‘seen’ \<(س)> and ‘sheen’ \<(ش)>. These two letters are composed of three successive strokes, with three dots above the second stroke in the case of ‘sheen’. A stroke is a PC that represents a part of a character having a one-pixel thickness. Because each stroke passes through the LMT, the ECC algorithm generates three cut indices, instead of one. In order to filter these cuts, we define a seen-stroke as a stroke with no dots above or below the baseline and no hole. A seen-stroke is also characterized by having a small peak above the baseline (i.e. is relatively short) and a flat structure near the baseline (does not dip below the baseline). Also, we define a sheen-stroke as a seen-stroke with dots above the baseline.

Based on the above definitions, the false cuts in the ‘seen’ or ‘sheen’ letters in the start and middle of the word \<(ـسـ، ـشـ)> will be filtered by detecting three successive seen-strokes for the ‘seen’ case or two seen-strokes with one sheen-stroke in between for the ‘sheen’ case. Nevertheless, this method filters the false cuts of the two letters in the start and the middle of the word, but it fails to detect false cuts when they appear at the end of the word.

We propose an improvement over this algorithm by taking into consideration the unique characteristic of the ‘seen’ and ‘sheen’ letters when they exist at the end of the word. This characteristic is the bowl shape seen at the end of these letters \<(ـس ، ـش)>. We define a bowl as (1) a PC with no dots above or below the baseline (2) a PC such that its right cut index must have at least one black pixel and its left cut index must contain no black pixels. This means that the bowl must be directly connected through the baseline to a PC before it and must not be connected to any PC after it. (3) a PC such that there must exist a region where the baseline vanishes within but resurfaces again (i.e. there exists a dip below the baseline). (4) a PC with a relatively small peak above the baseline. 

Based on this improved algorithm, the false cuts (the first two cut indices) in the ‘seen’ letter will be filtered when the IFC algorithm detects three successive seen-strokes or two successive seen-strokes followed by a bowl, whereas the false cuts (the first two cut indices) in the ‘sheen’ letter will be filtered when the IFC algorithm detects two seen-strokes separated by a sheen-stroke or a seen-stroke followed by a sheen-stroke and a bowl.


\textbf{ii. ‘Saad’ and ‘Daad’ case \<(ص، ض، صـ، ضـ)>:} The second set of characters that result in an additional false cut is ‘saad’ and ‘daad’ \<(ص، ض، ـصـ، ـضـ)>. To filter the false cuts in these two letters, we define a hole as a PC containing a rounded area of white pixels (a hole) enclosed by a larger rounded area of black pixels. The two letters consist of a hole followed by a seen-stroke when they appear at the start or the middle of a word, or a hole followed by a bowl when they appear at the end of a word. Initially, we wrote our cut filtration algorithm so that whenever it encountered such a case, it merged the two PCs into one by removing the cut index in-between. 

However, an interesting case that generated confusion with this definition was a ‘meem’ or a ‘faa’ followed by a ‘daal’ \<(مد ، فـد)>. Both cases would always be misinterpreted as a ‘saad’ or ‘daad’ and the cut filtration algorithm would falsely merge the ‘daal’ and the preceding character into one.

Therefore, we defined a saad-stroke to differentiate between the ‘daal’ stroke and the strokes of ‘saad’ and ‘daad’. The saad-stroke is a seen-stroke with the additional condition of being surrounded by cut indices that have at least one black pixel each. The goal of this extra specification is to ensure that the saad-stroke is connected to the baseline from both sides and is not followed by a space, as is the case with ‘daal’. As such, the false cuts in ‘saad’ and ‘daad’ can be filtered when the cut filtration algorithm finds a hole followed by a saad-stroke or when it finds a hole followed by a bowl. 

\textbf{iii. ‘Baa’, ‘Taa’, ‘Thaa’ and ‘Faa’ case \<(ـب، ــت، ــث، ــف)>:} There is also a difficult case where a ‘baa’, ‘taa’, ‘thaa’ or ‘faa’ \<(ـب، ــت، ــث، ــف)> at the end of a word may cause a false cut. This occurs when the stroke at the end of the aforementioned characters is tall enough to intersect with the LMT. In this case, the ECC algorithm will generate a false cut at this stroke position which will need to be filtered.

In order to filter this extra false cut, we define an end stroke as a ‘seen’ stroke that has additional restrictions. Firstly, an end stroke must be followed by a cut index that does not intersect the baseline and preceded by a cut index that intersects the baseline. Furthermore, we locate the leftmost and the uppermost black pixels of the PC and we calculate the horizontal distance d between these two pixels. If d is measured to be less than or equal to 2 pixels, the ICF algorithm identifies the PC as an end stroke and removes the preceding cut. 

It is worth noting that the extra restriction on the horizontal distance between the top-leftmost and the uppermost black pixels is essential in order not to incorrectly identify the ‘daal’ letter \<(ــد)> as an end stroke because the stroke in the ‘daal’ letter has geometrical features that resemble the second stroke in the ‘taa’ and ‘thaa’ letters. 
Algorithm 2. ICF Algorithm

\begin{algorithm}
\caption{ICF Algorithm}
\label{ICF_Algorithm}
\begin{algorithmic}[1]

\State \textbf{Input: $PCArray$}
\State {$FilteredCharacterArray \gets \emptyset$}

\ForAll {$PC$ in $PCArray$} 
	\If {$PC$ is $Seen Str$}
		\If {$NPC$ is $Seen Str$ or $NPC$ is $Sheen Str$}
			\If {$ANPC$ is $Seen Str$ or $Bowl$}
				\State $PCArray.Merge(PC, NPC, ANPC)$
			\EndIf
		\EndIf
	\EndIf
\EndFor

\ForAll {$PC$ in $PCArray$} 
	\If {$PC$ is $Saad Str$ or $PC$ is $Bowl$}
		\State $PCArray.Merge(PreviousPC, PC)$
	\EndIf
\EndFor

\ForAll {$PC$ in $PCArray$} 
	\If {$PC$ is $End Str$ with $D <= 2$}
		\State $PCArray.Merge(PreviousPC, PC)$	
	\EndIf
\EndFor
\State $FilteredCharacterArray \gets PCArray$
\State \textbf{Output: $FilteredCharacterArray$}

\end{algorithmic}
\end{algorithm}

Although the previous cut filtration cases may seem largely independent of each other, the filtration order can greatly affect the character segmentation performance. For instance, executing the saad/daad-case before the seen/sheen-case may cause the algorithm to confuse the first stroke of the ‘seen’ letter as a ‘saad’ or a ‘daad’ stroke, and falsely merge it with the preceding character. As shown in Algorithm \ref{ICF_Algorithm}, our ICF algorithm filters all PCs according to the order of filtration cases mentioned in this paper.

It is worth mentioning that our character segmentation algorithm does not depend on any linguistic or statistical patterns in the Arabic language and is well-equipped to segment any sequence of Arabic letters. The previous work of \cite{qaroush2019efficient} relied on their cut filtration algorithm’s assumption that some letters such as ‘saad’ and ‘daad’ are never followed by ‘seen’ or ‘sheen’. On the other hand, our ICF algorithm is general enough to segment any Arabic word regardless of the arrangement of the letters in the word.

Our character segmentation algorithm provides several improvements over the work of \cite{qaroush2019efficient}. As opposed to the algorithm proposed by \cite{qaroush2019efficient}, the ECC algorithm does not generate a cut at positions where there are black pixels above or below the baseline, and this improvement reduces the number of false cuts that the cut filtration algorithm has to filter. The second improvement in our ICF is filtering each PC based on clear and specific structural features of Arabic letters such as ‘seen-stroke’, ‘sheen-stroke’, ‘saad-stroke’, ‘end-stroke’, ‘bowl’ and ‘hole’. These features are essential for the ICF algorithm in order to not accidentally remove any valid cut. Previous work in character segmentation \cite{qaroush2019efficient} did not provide a correct exact definition of a bowl and, as a result, the characters having a bowl-shape in their structure such as ‘noon’, ‘qaaf’, ‘yaa’, ‘raa’, and ‘zaay’ letters \<(ن، ق، ي، ر، ز)> were confused with the bowl part of the ‘saad’ and ‘daad’ letters \<(ص، ض)> and were falsely merged with the preceding character.

The third improvement is filtering the false cuts in the case of the letters ‘baa’, ‘taa’, ‘thaa’, and ‘faa’ based on the leftmost black pixel and the uppermost black pixel instead of the top-leftmost black pixel only. The aforementioned false cuts are detected in \cite{qaroush2019efficient}’s algorithm if the top-leftmost black pixel has a relatively small height when compared to the highest black pixel in the line. As a result, almost all letters that occur at the end of a word will be falsely merged with the preceding character. Only the ‘alef’ \<( أ )> character will not be falsely merged when written at the end of a word since its top-leftmost black pixel is relatively tall, while every other character in the Arabic alphabet will have a relatively low top-leftmost black pixel when written at the end of a word. 

Finally, our segmentation algorithm solves the challenging case of the ‘seen’ \<( س )> and ‘sheen’ \<( ش )> letters at the end of the word, which was not handled by the algorithm proposed by \cite{qaroush2019efficient} and resulted in a considerable number of incorrect segmentations. We conclude that our algorithm improves on the one proposed by Qaroush \textit{et al.} \cite{qaroush2019efficient} and addresses many of its shortcomings in the character segmentation problem.
\subsection{Character Recognition Model}
We propose feeding the images of the segmented characters to an artificial neural network since this will yield better performance than choosing features manually. The architecture of the ANN used in this research is a multilayered feed-forward network architecture with four layers. This neural network learns highly complex nonlinear features by training on a sufficiently large training set of Arabic characters together with their ground-truth labels.
   
We will generate our own training set by comparing the number of segmented letters generated from a word-using our proposed algorithms- with the number of letters in the ground truth word. If the numbers match, we associate each letter with its corresponding label. If the numbers do not match, then we skip and discard this word. It is worth noting that this over/under-segmentation problem arises if the scanned document was too noisy or blurry. However, we are cautious about the training set and prefer not to risk accidentally training with false characters.

\begin{figure}
\centering
\includegraphics[width=3in]{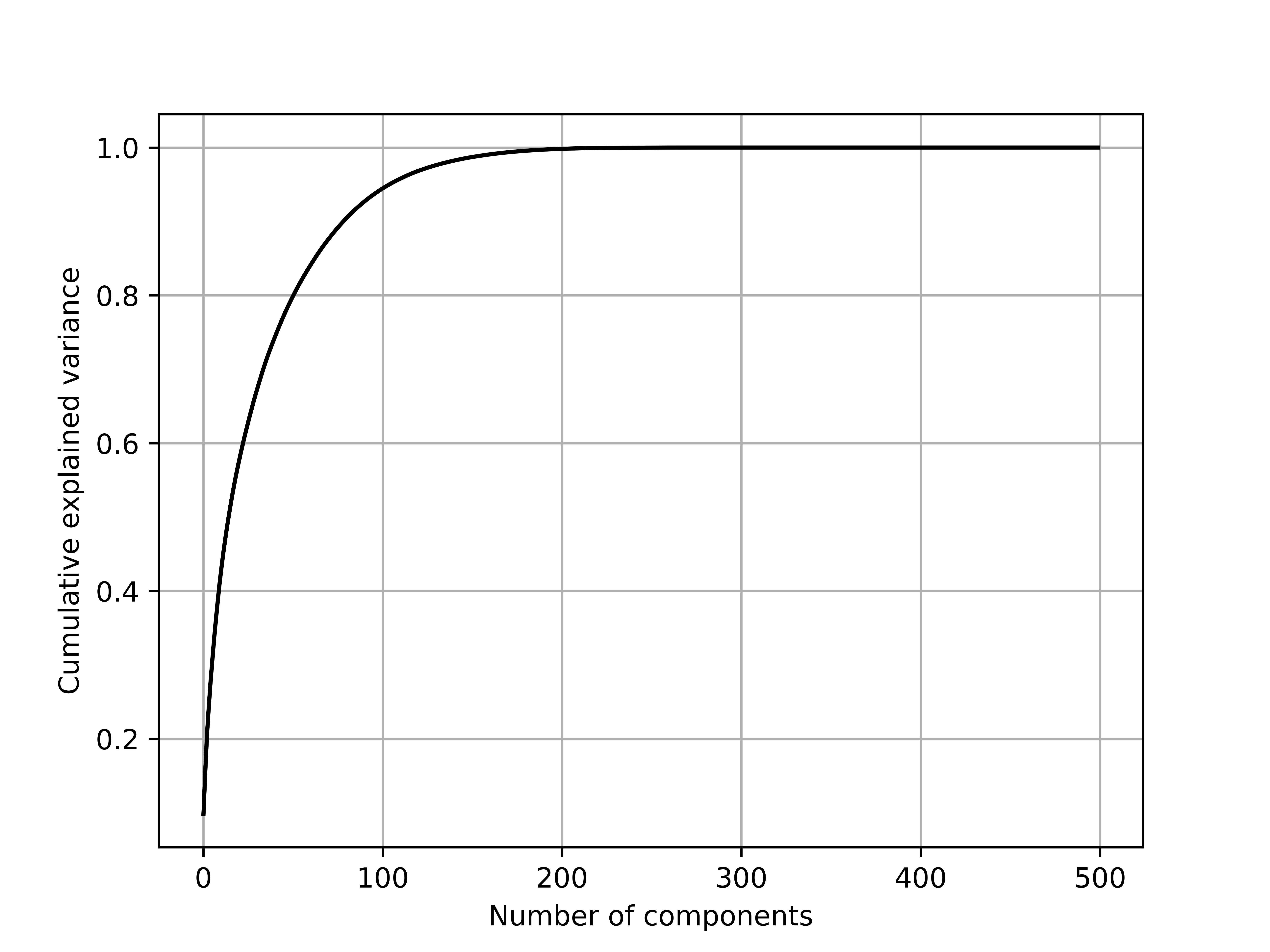}
\caption{PCA Dimensionality Graph}
\label{PCA}
\end{figure}

We begin by resizing all the images generated from the character segmentation algorithm to be 24x24 pixels and then flattening them to be 576-dimensional vectors. In addition, we perform dimensionality reduction on these 576-dimensional vectors using Incremental Principal Component Analysis (IPCA), which -as illustrated in Figure \ref{PCA} can represent the original vectors using only 200 principal components while retaining 99\% of the total variance in the data. These 200-dimensional vectors are then fed to the neural network that consists of two hidden layers and a softmax output layer of sizes 150, 70, and 29 respectively. The softmax output layer assigns a likelihood value for each character of the 29 characters. This value represents the probability that this character is the correct classification for the input image. The inference phase of the model then classifies the input character as the character with the highest likelihood of being the correct prediction.

We trained the model with our generated dataset that consists of 1,200,000 images of perfectly segmented Arabic letters in Naskh script using mini-batch training of batch size 8192 for 80 epochs. Categorical cross-entropy loss is calculated in the output layer and optimized using Adam optimizer \cite{kingma2014adam} with hyper-parameters $\beta_1$ and $\beta_2$ being 0.9 and 0.999 respectively. For the hidden layers, the Rectified Linear Unit (ReLU) is used as a non-linear activation function followed by a 10\% dropout layer to ensure elimination of overfitting as much as possible. A learning rate of 0.001 with no decay factor is used. The network is initialized using He initialization \cite{he2015delving} and shuffled per epoch. We used a validation set of 12,000 images, representing 1\% of the training set.
\subsection{Post-processing}
The character recognition model outputs a predicted class for each of the character images. However, a final step of post-processing is necessary to aggregate these characters into words and separate them with spaces to generate a meaningful text document. The previously predicted letters are produced consecutively with no spaces until we encounter an activated End of Word (EOW) flag. Every character has an EOW flag value; values for this flag are obtained in the character segmentation phase by setting the EOW flag to true (activated) for the final segment of the word that is being segmented. In the post-processing step, whenever we encounter a character with an activated EOW flag, indicating the end of a word, we place a space directly after this character. This step ensures the production of a comprehensible document.
\section{Experimental Results}
\subsection{Datasets}
There are several datasets for Arabic character recognition \cite{lawgali2013hacdb}. We used the open dataset Watan-2004 corpus \cite{watandataset} for training our neural network. The dataset includes very large Arabic vocabulary, with different font types, sizes, and styles and is rich with separable characters, overlapping characters, and ligatures. The scanned images in the dataset are 72 dpi resolution images. For training the model, we randomly selected 550 documents/images of plain Arabic Naskh from different topics, approximately 282,000 words, or 1,200,000 characters. For validation and testing, we randomly chose 6 and 10 documents/images of Arabic Naskh script as a validation set and a test set respectively, of sizes 3300 words (12000 characters) and 5500 words (100,500 characters).

We further experimented with the system with different test sets of plain Arabic fonts; Naskh, Transparent Arabic (T. Arabic), Simplified Arabic (S. Arabic), M. Unicode Sara, Tahoma, Times New Roman, and Arial with different font sizes; 10, 12, 14, and 16. We tested the system on the APTI dataset (Arabic Printed Text Image) \cite{slimane2009new}, which is a standard benchmarking dataset for Arabic OCR tasks. We used Keras \cite{chollet2015keras} for training our model and we ran our experiments on a core i7 5820K 3.3 GHz machine, with 32 GB RAM, Ubuntu OS 16.04, and GPU NVIDIA RTX 2070 with 8 GB memory and 2560 cores.
\subsection{Results and Evaluation}	
This section evaluates our word segmentation algorithm, character segmentation algorithm, character recognition model, as well as the overall system accuracy. We used both Watan-2004 and a subset of the APTI datasets for evaluating our system. We also compare our segmentation algorithms to the work of Anwar \textit{et al.} \cite{anwar2015segmentation}, Radwan \textit{et al.} \cite{radwan2016predictive}, and Mousa \textit{et al.} \cite{mousa2017arabic}. We compare the overall system performance when our improved segmentation algorithm is used against when the segmentation algorithm of Qaroush \textit{et al.} \cite{qaroush2019efficient} is used.

\subsubsection{Word Segmentation}
We define the word segmentation accuracy as the number of correctly segmented words divided by the total number of actual words in the document. Our method achieves an average word segmentation accuracy of 99.94\% for different fonts, as indicated in Table \ref{Table1} outperforming Qaroush \textit{et al.} \cite{qaroush2019efficient} for different fonts for word segmentation.

\begin{table}
\renewcommand{\arraystretch}{1.3}
\caption{Word Segmentation Comparison with Qaroush \textit{et al.} \cite{qaroush2019efficient}}
\label{Table1}
\centering
\begin{tabularx}{\linewidth}{|L|L|L|L|L|}
\hline

\multicolumn{1}{|c|}{\multirow{2}{*}{\textbf{Font}}} & \multicolumn{2}{c|}{\textbf{Method Proposed}} & \multicolumn{2}{c|}{\textbf{Qaroush \textit{et al.} }\cite{qaroush2019efficient}}\\
\cline{2-5}
& Words & Accuracy & Words & Accuracy\\
\hline

Tahoma & 25,928 &\textbf{99.96\%} & 2,319 & 99.4\%\\
\hline

Naskh & 29,169 &\textbf{99.95\%} & 2,921 & 96.1\%\\
\hline

S. Arabic & 22,687 &\textbf{99.94\%} & 2,884 & 98.8\%\\
\hline

T. Arabic & 21,055 &\textbf{99.90\% }& 2,860 & 99.1\%\\
\hline

\end{tabularx}
\end{table}

\subsubsection{Character Segmentation}
We first segment each word and compare the number of segmented characters with the number of actual characters in the word. Then we count their absolute difference as incorrectly segmented characters. We finally define the character segmentation accuracy as the total number of correctly segmented characters divided by the total number of actual characters. Our method achieves an average character segmentation accuracy of 98.23\% for different fonts, as indicated in Table \ref{Table2}, where we also compare our results for each font with \cite{qaroush2019efficient}. Comparing our results with the results of \cite{anwar2015segmentation} and \cite{mousa2017arabic}, as shown in Table \ref{Table3}, we note that our system outperforms all other character segmentation algorithms in the character segmentation task.

\begin{table}
\renewcommand{\arraystretch}{1.3}
\caption{Character Segmentation Comparison with Qaroush \textit{et al.} \cite{qaroush2019efficient}}
\label{Table2}
\centering
\begin{tabularx}{\linewidth}{|L|L|L|L|L|}
\hline

\multicolumn{1}{|c|}{\multirow{2}{*}{\textbf{Font}}} & \multicolumn{2}{c|}{\textbf{Method Proposed}} & \multicolumn{2}{c|}{\textbf{Qaroush \textit{et al.}}\cite{qaroush2019efficient}}\\
\cline{2-5}
& Characters & Accuracy & Characters & Accuracy\\
\hline

Tahoma & 114,080 & \textbf{97.80\%} & 12,262 & 97.00\%\\
\hline

Naskh & 131,260 & \textbf{98.66\%} & 12,585 & 94.52\%\\
\hline

S. Arabic & 100,957 & \textbf{99.06\%} & 13,572 & 96.10\%\\
\hline

T. Arabic & 89,483 & \textbf{97.40\%} & 13,120 & 96.26\%\\
\hline

\end{tabularx}
\end{table}

\begin{table*}
\renewcommand{\arraystretch}{1.5}
\caption{Character Segmentation Results}
\label{Table3}
\centering
\begin{threeparttable}
\begin{tabularx}{\linewidth}{|L|L|L|L|L|L|}
\hline

\textbf{Method} & \textbf{Dataset} & \textbf{Character Count} & \textbf{Font Size} & \textbf{Font Style} & \textbf{Accuracy}\\
\hline

Anwar \textit{et al.} \cite{anwar2015segmentation} & Self-generated & Not Reported & 70pt & Traditional Arabic & 97.55\%\\
\hline

Mousa \textit{et al.} \cite{mousa2017arabic} & Self-generated & Not Reported & Not Reported & Not reported & 98.00\%\\
\hline

Qaroush \textit{et al.} \cite{qaroush2019efficient} & APTI \cite{slimane2009new} & 100K characters & 10, 12, 14, 16, 18, and 24 & Font Set A\tnote{\(\dagger\)} & 97.50\%\\
\hline

Method Proposed & Watan-2004 \cite{watandataset} + APTI \cite{slimane2009new} & 1M characters & 10, 12, 14, and 16 & Font Set B\tnote{\(\ddagger\)} & \textbf{98.23}\%\\
\hline

\end{tabularx}
\begin{tablenotes}
    \item[ \(\dagger\) ] \textbf{Font Set A:} Naskh, T. Arabic, S. Arabic, M Unicode Sara, Tahoma and Advertising Bold.
    \item[ \(\ddagger\) ] \textbf{Font Set B:} Font Set A, Andalus, Diwani, Thuluth
\end{tablenotes}
\end{threeparttable}
\end{table*}

\subsubsection{Overall System Performance}
The neural network achieved average recognition accuracy of 99.89\%. The overall system accuracy is measured by calculating the Levenshtein edit distance between the generated document and the actual document. We show that our system achieves an overall accuracy of 97.94\%. Therefore, and to the best of our knowledge, we propose that our Arabic OCR system is superior to all other segmentation based Arabic OCR in terms of accuracy and running time. It is also worth mentioning that our proposed system was developed and trained using a very large dataset, which is rich with Arabic text in different font types and sizes. Table 4 shows the evaluation of our system in comparison to the work of Touj \textit{et al.} \cite{touj2005generalized} as well as the segmentation method of Qaroush \textit{et al.}\cite{qaroush2019efficient} followed by an ANN for recognition.

\begin{table}
\renewcommand{\arraystretch}{1.5}
\caption{Overall System Evaluation}
\label{Table4}
\centering
\begin{tabularx}{\linewidth}{|L|L|L|L|L|}
\hline

\textbf{Method} & \textbf{Dataset} & \textbf{Word Count} & \textbf{System Accuracy} & \textbf{Avg. Run Time / 550 words}\\
\hline

Touj \textit{et al.} \cite{touj2005generalized} & Not Reported & 1500 & 97.00\% & Not reported\\
\hline

Qaroush \textit{et al. }\cite{qaroush2019efficient} + ANN & Watan-2004 \cite{watandataset} & 3300  & 94.95\% & 3.42 seconds\\
\hline

Method Proposed & Watan-2004 \cite{watandataset}, APTI \cite{slimane2009new} & 3300 & \textbf{97.94}\% & \textbf{1.49 seconds}\\
\hline

\end{tabularx}
\end{table} 

\section{Future Work}
Much like other OCR systems, our system’s performance decreases when operating on documents with high noise. This creates the demand to work on more elaborate preprocessing methods in the future without sabotaging our system’s remarkable running time. Also, our recognition model is currently limited to the fonts that it inferred from the training set; as a result of that, we aim to enrich the training set to capture more fonts and possibly all fonts without ligatures.

Apart from the low-level additions to our system, potential work on this system would include integrating it into larger applications such as image-speech systems. Such applications, where instant results are crucial, will be a perfect fit for our method because of the very efficient running time we provide.

\section{Conclusion}
Arabic Optical Character Recognition introduces many challenges in the character segmentation and recognition phase. This paper proposes a complete language-independent Arabic OCR pipeline with an improved character segmentation algorithm based on word-level features and a bio-inspired character recognition model based on neural networks. We proposed a highly accurate and efficient system. The overall system architecture consists of: a simple yet effective pre-processing module, an enhanced reliable module for character segmentation, an artificial neural network for the recognition of segmented characters, and finally a post-processing module that formulates the output of our system into a digitized document. We evaluated our system on different datasets with high variability in font types and sizes. The experimental results show that our system outperforms the current state-of-the-art algorithms for word segmentation, character segmentation, and character recognition. We evaluated the overall performance of the system and concluded that our system achieves outstanding results in accuracy and running time.
\bibliographystyle{IEEEtran}
\bibliography{ms}
\end{document}